%% file: MPRSS2022.tex
\documentclass[runningheads]{llncs}

\usepackage{graphicx}
\usepackage{cite}
\usepackage{amsfonts}
\usepackage{algorithmic}
\usepackage{textcomp}
\usepackage{xcolor}
\usepackage{hyperref}
\usepackage{url}
\usepackage{times}

\usepackage{soul}
\usepackage[utf8]{inputenc}
\usepackage[list=true]{subcaption}
\usepackage{graphicx}
\usepackage{amsmath}
\usepackage{amssymb}
\usepackage{booktabs}
\usepackage{tabu}
\usepackage{wrapfig}
\usepackage{arydshln}
\usepackage[titletoc]{appendix}
\usepackage{bibunits}
\usepackage[textsize=tiny]{todonotes}
\usepackage[normalem]{ulem}
\usepackage{multirow}
\usepackage{siunitx}

\newcommand\freefootnote[1]{
  \let\thefootnote\relax
  \footnotetext{#1}
  \let\thefootnote\svthefootnote
}

\begin{document}

\title{Representation Learning for Tablet and Paper Domain Adaptation in Favor of Online Handwriting Recognition}
\titlerunning{Tablet and Paper Domain Adaptation for Online Handwriting Recognition}

\input{00authors.tex}
\maketitle
\textbf{Copyright held by Owner/Author. This is the author's version of the work. It is posted here for your personal use. Not for distribution. The definite Version of Record was accepted at \textit{IAPR Intl. Workshop on Multimodal Pattern Recognition of Social Signals in Human Computer Interaction (MPRSS)}, August 2022.}
\input{00abstract.tex}
\input{01introduction.tex}
\input{02related_work.tex}
\input{03method.tex}
\input{04experiments.tex}
\input{05evaluation.tex}
\input{06conclusion.tex}

\bibliographystyle{splncs04}
\bibliography{MPRSS2022}

\end{document}

%% file: 00authors.tex
\author{\underline{Felix Ott}\inst{1,2}\orcidID{0000-0002-4392-0830} \and
David Rügamer\inst{2,3}\orcidID{0000-0002-8772-9202} \and \\
Lucas Heublein\inst{1}\orcidID{0000-0001-6670-3698} \and
Bernd Bischl\inst{2}\orcidID{0000-0001-6002-6980} \and
Christopher Mutschler\inst{1}\orcidID{0000-0001-8108-0230}}

\authorrunning{Ott et al.}

\institute{Fraunhofer IIS, Fraunhofer Institute for Integrated Circuits IIS \\
\email{\{felix.ott, heublels, christopher.mutschler\}@iis.fraunhofer.de} \and
LMU Munich, Munich, Germany \\
\email{\{david.ruegamer, bernd.bischl\}@stat.uni-muenchen.de} \and
RWTH Aachen, Aachen, Germany}

%% file: 00abstract.tex
\begin{abstract}
The performance of a machine learning model degrades when it is applied to data from a similar but different domain than the data it has initially been trained on. The goal of domain adaptation (DA) is to mitigate this domain shift problem by searching for an optimal feature transformation to learn a domain-invariant representation. Such a domain shift can appear in handwriting recognition (HWR) applications where the motion pattern of the hand and with that the motion pattern of the pen is different for writing on paper and on tablet. This becomes visible in the sensor data for online handwriting (OnHW) from pens with integrated inertial measurement units. This paper proposes a supervised DA approach to enhance learning for OnHW recognition between tablet and paper data. Our method exploits loss functions such as maximum mean discrepancy and correlation alignment to learn a domain-invariant feature representation (i.e., similar covariances between tablet and paper features). We use a triplet loss that takes negative samples of the auxiliary domain (i.e., paper samples) to increase the amount of samples of the tablet dataset. We conduct an evaluation on novel sequence-based OnHW datasets (i.e., words) and show an improvement on the paper domain with an early fusion strategy by using pairwise learning.\freefootnote{Supported by the Federal Ministry of Education and Research (BMBF) of Germany by Grant No. 01IS18036A and by the project ``Schreibtrainer'', Grant No. 16SV8228, as well as by the Bavarian Ministry for Economic Affairs, Infrastructure, Transport and Technology through the Center for Analytics-Data-Applications within the framework of ``BAYERN DIGITAL II''.}

\keywords{Online handwriting recognition (OnHW) \and sensor pen \and domain adaptation (DA) \and deep metric learning (DML) \and writer-(in)dependent tasks.}
\end{abstract}

%% file: 01introduction.tex
 \section{Introduction}
\label{chap_introduction}

\begin{figure}[b!]
	\centering
	\begin{minipage}[b]{0.48\linewidth}
        \centering
        \includegraphics[trim=12 10 8 8, clip, width=1.0\linewidth]{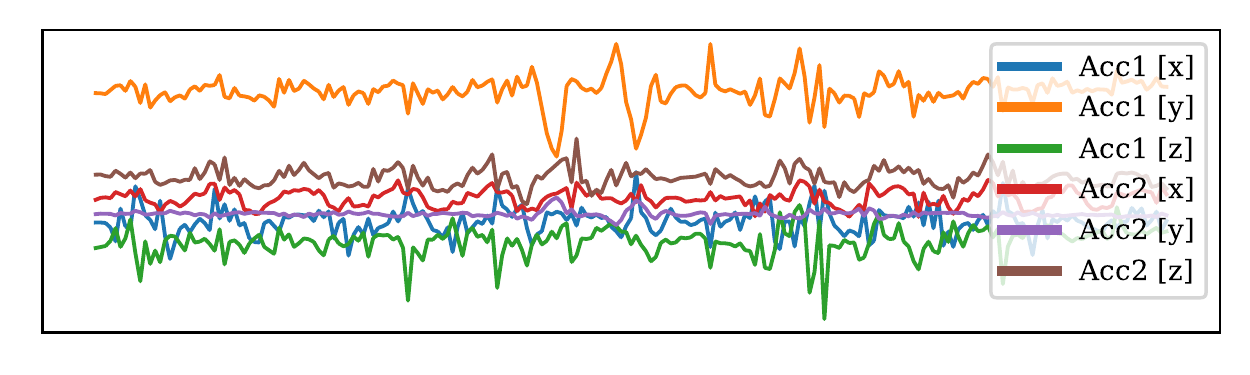}
    \end{minipage}
    \hfill
    \begin{minipage}[b]{0.48\linewidth}
        \centering
        \includegraphics[trim=12 10 8 8, clip, width=1.0\linewidth]{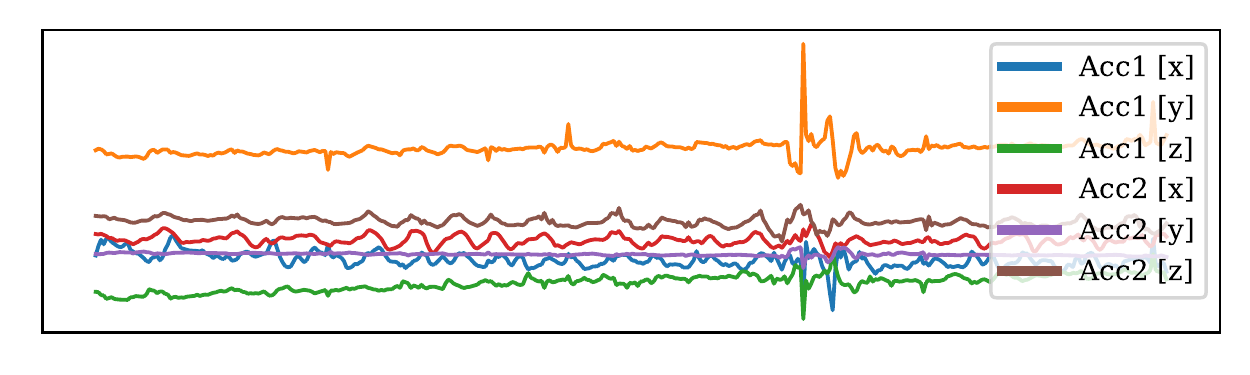}
    \end{minipage}
    \begin{minipage}[b]{0.48\linewidth}
        \centering
        \includegraphics[trim=12 10 8 8, clip, width=1.0\linewidth]{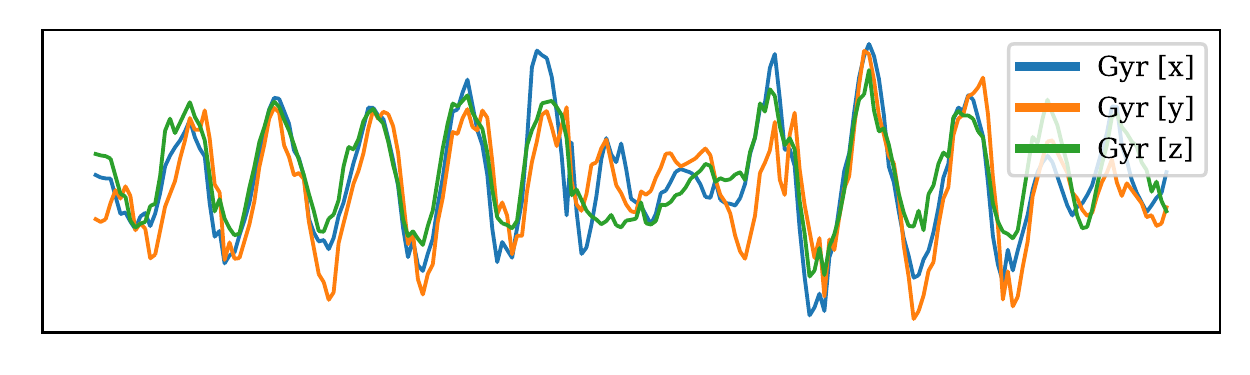}
    \end{minipage}
    \hfill
    \begin{minipage}[b]{0.48\linewidth}
        \centering
        \includegraphics[trim=12 10 8 8, clip, width=1.0\linewidth]{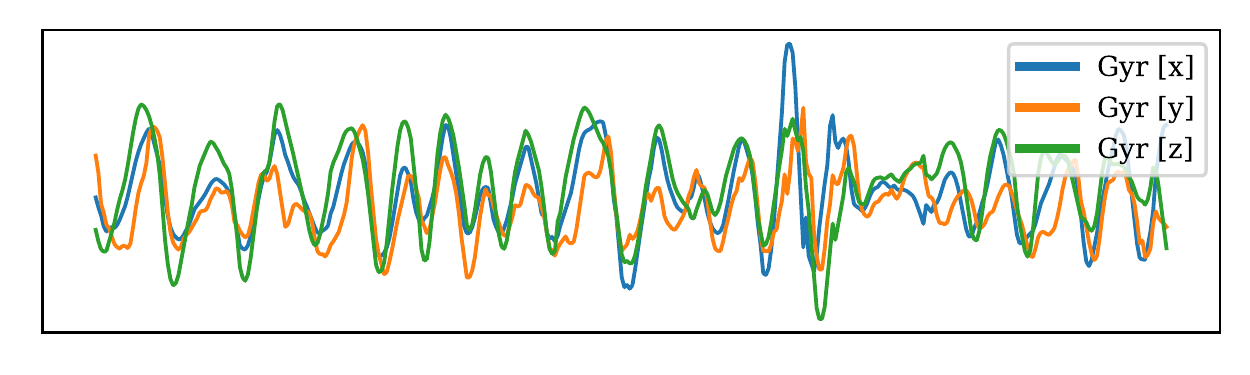}
    \end{minipage}
    \begin{minipage}[b]{0.48\linewidth}
        \centering
        \includegraphics[trim=12 10 8 8, clip, width=1.0\linewidth]{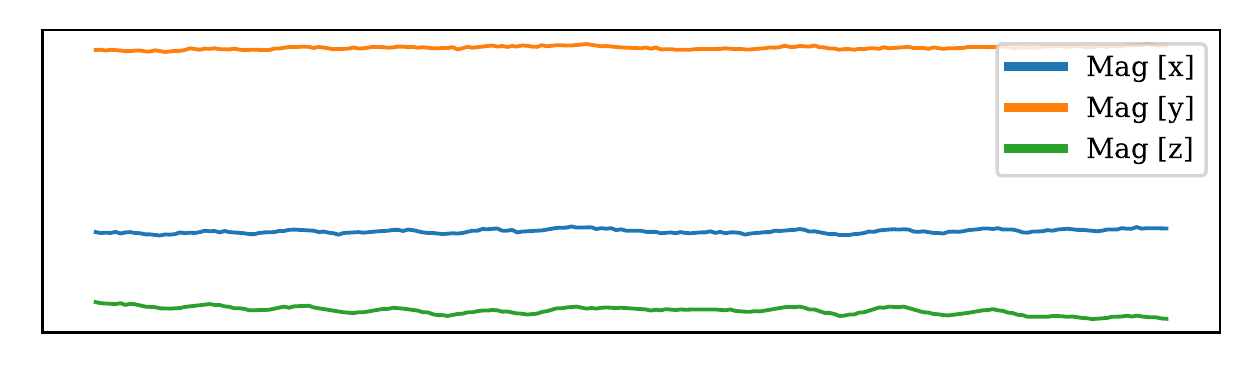}
    \end{minipage}
    \hfill
    \begin{minipage}[b]{0.48\linewidth}
        \centering
        \includegraphics[trim=12 10 8 8, clip, width=1.0\linewidth]{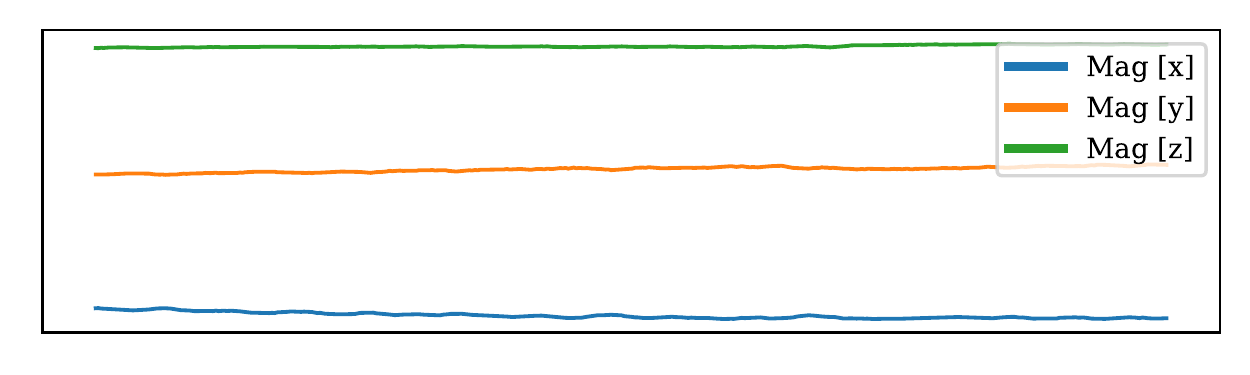}
    \end{minipage}
    \begin{minipage}[b]{0.48\linewidth}
        \centering
        \includegraphics[trim=12 2 8 8, clip, width=1.0\linewidth]{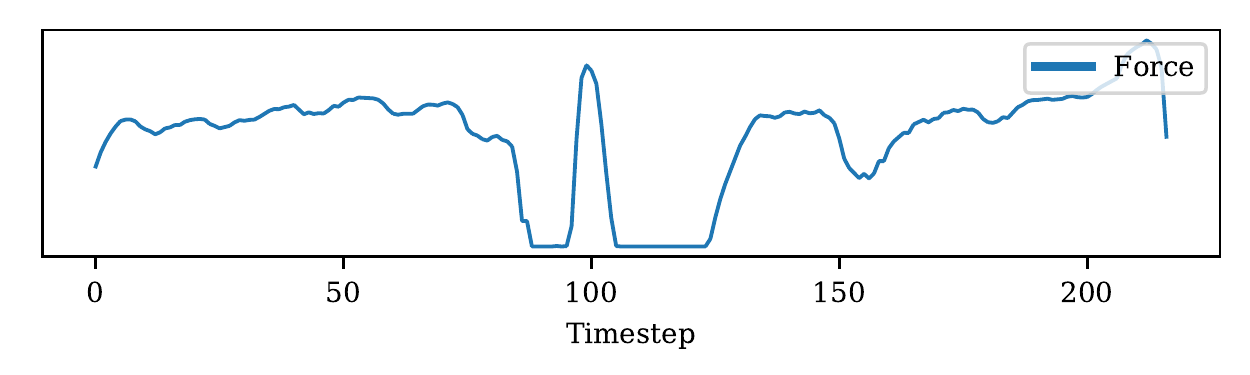}
    \end{minipage}
    \hfill
    \begin{minipage}[b]{0.48\linewidth}
        \centering
        \includegraphics[trim=12 2 8 8, clip, width=1.0\linewidth]{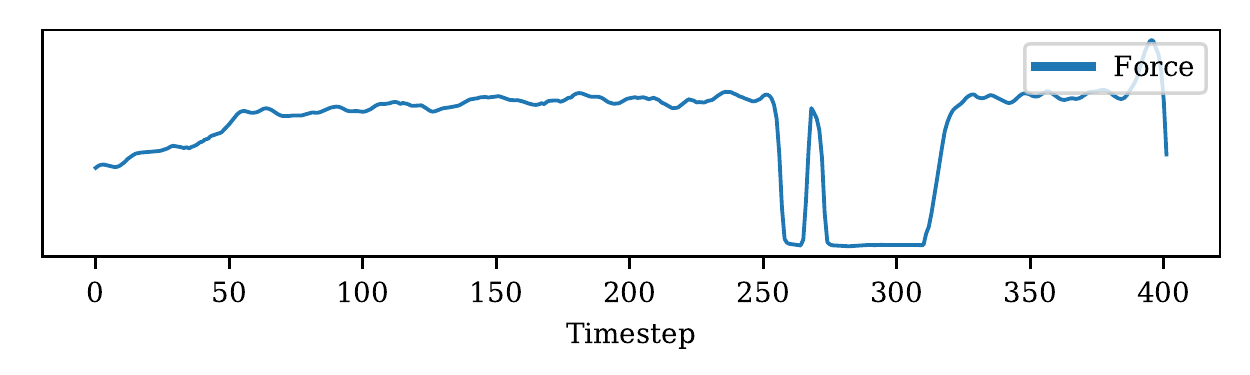}
    \end{minipage}
	\begin{minipage}[b]{0.48\linewidth}
        \centering
        \includegraphics[trim=0 100 0 1000, clip, width=1.0\linewidth]{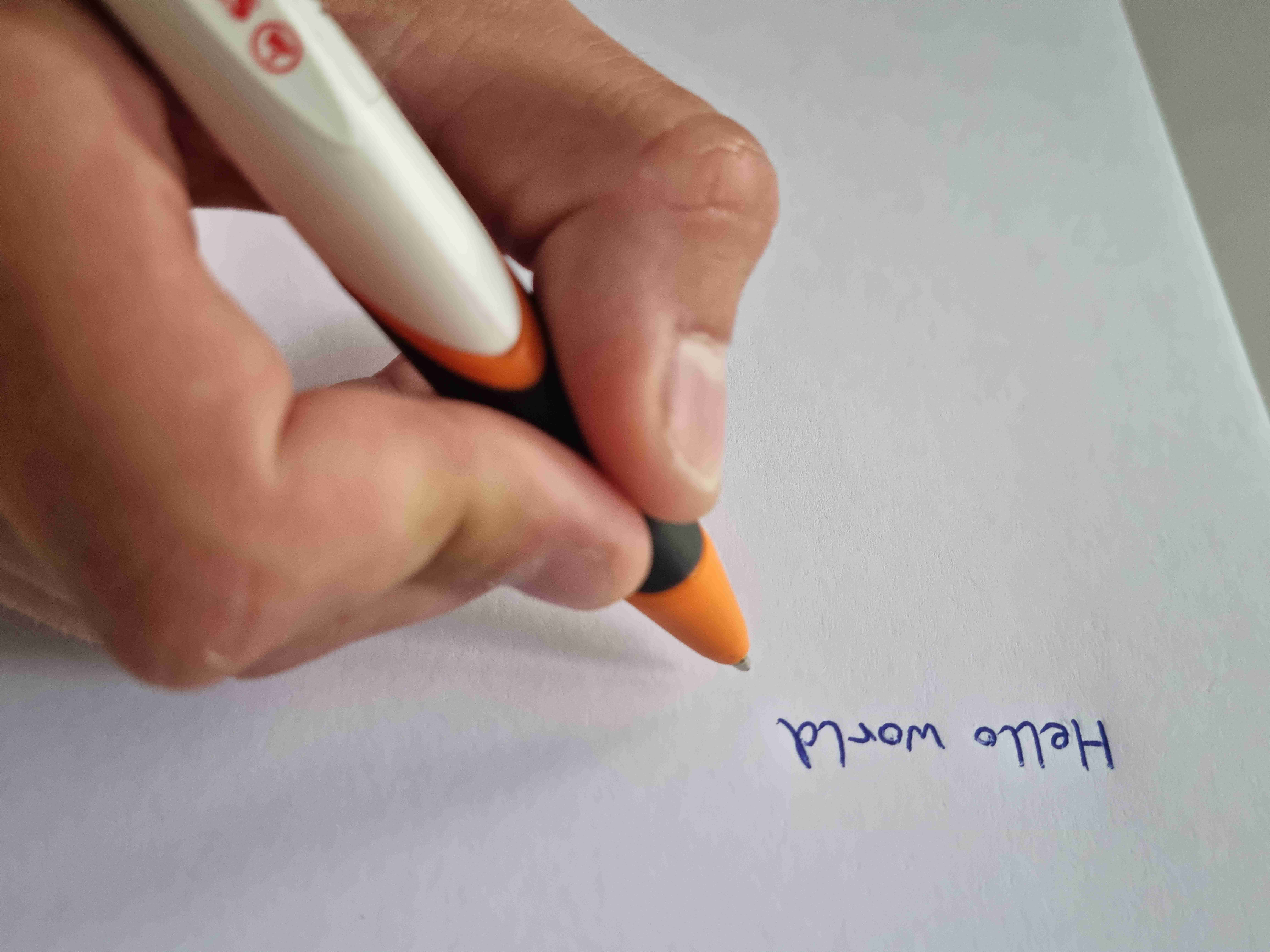}
    \end{minipage}
    \hfill
	\begin{minipage}[b]{0.48\linewidth}
        \centering
        \includegraphics[trim=0 100 0 1000, clip, width=1.0\linewidth]{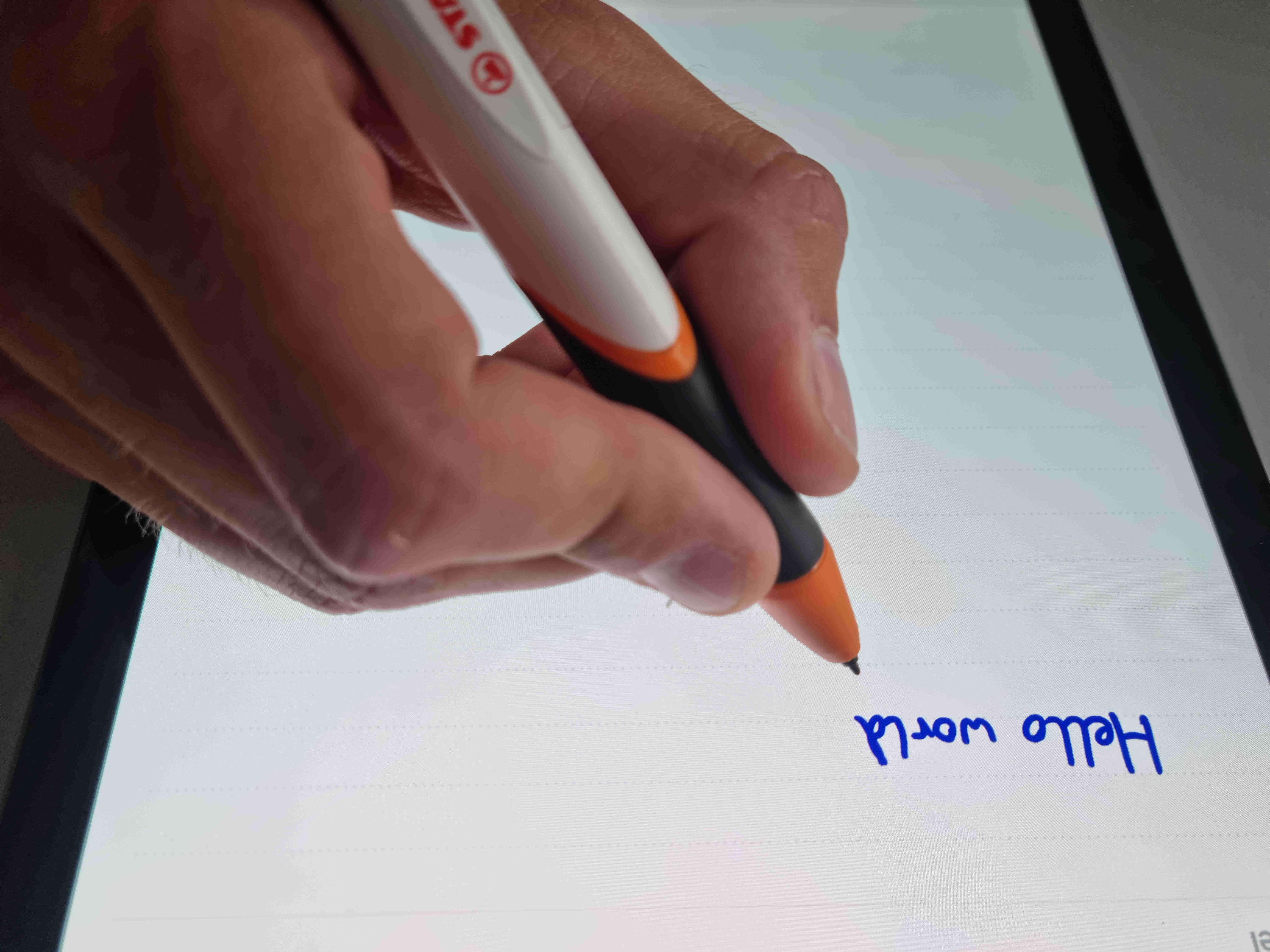}
    \end{minipage}
    \caption{Comparison of accelerometer (1\textsuperscript{st} row), gyroscope (2\textsuperscript{nd} row), magnetometer (3\textsuperscript{rd} row) and force (4\textsuperscript{th} row) data from a sensor pen \cite{ott} on paper (left) and tablet (right).}
    \label{image_teasure_figure}
\end{figure}

HWR can be categorized into offline and online HWR. While offline HWR deals with the analysis of the visual representation, OnHW recognition works on different types of spatio-temporal signals and can make use of temporal information such as writing direction and speed \cite{plamondon}. Typically, recording systems make use of a stylus pen together with a touch screen surface \cite{alimoglu}. Systems for writing on paper became popular, first prototypical systems were used \cite{deselaers}, and recently a novel system enhanced with inertial measurement units (IMUs) became prominant \cite{ott}. These IMU-enhanced pens are real-world applicable. While previous work~\cite{ott, ott_tist, ott_mm, ott_ijcai, klass_lorenz} used this pen for writing on paper, \cite{ott_wacv} used this pen for writing on tablet. Figure~\ref{image_teasure_figure} presents IMU data from a sensor-enhanced pen for writing on paper (left) and tablet (right). Due to the rough paper, the sensor data for writing on paper has more noise than writing on surface. Furthermore, the magnetic field of the tablet influences the magnetometer of the pen. This leads to different distributions of data and a domain shift between both data sources. Previously, tablet and paper data are processed separately, and hence, there is no method that can use both data sources simultaneously and inter-changeably. 

Traditional ML algorithms assume training and test datasets to be \textit{independent and identically distributed}. When applied in practice a domain shift appears in test data (here, shift between sensor data from tablet and paper), and hence, this assumption rarely holds in practice \cite{sun_feng}. DA~\cite{tzeng, long_wang} tries to compensate for this domain shift by transferring knowledge between both data sources. Most techniques transform the source data (here, data written on paper) by minimizing the distance to the target data (here, data written on tablet) \cite{sun_feng}, or by transforming the extracted features of the data sources \cite{ott_mm}. To transform features of source domain into the target domain or to compare feature embeddings, higher-order moment matching (HoMM) \cite{chen_fu} is often employed. Typically, the maximum mean discrepancy (MMD) \cite{long_cao} (HoMM of order 1) or the kernelized method kMMD \cite{long_zhu_mmd} between features is evaluated. Correlation alignment (CORAL) \cite{sun_saenko} is of order 2. A related, yet different, task is pairwise learning. The \textit{pairwise contrastive loss} \cite{chopra} minimizes the distance between feature embedding pairs of the same class and maximizes the distance between feature embeddings of different classes dependent on a margin parameter. The \textit{triplet loss} \cite{schroff} defines an anchor and a positive as well as a negative point, and forces the positive pair distance to be smaller than the negative pair distance by a certain margin. While the triplet loss is typically used for image recognition, \cite{ott_ijcai, guo_tang} used this loss for sequence-based learning. 

\textbf{Contributions.} We propose a method for OnHW recognition from sensor-enhanced pens for classifying words written on tablet and paper. We address the task of representation learning of features from different domains (i.e., tablet and paper) by using moment matching techniques (i.e., MMD and CORAL). For matching positive and negative samples of paper datasets w.r.t. anchor samples of tablet datasets, we use a triplet loss with dynamic margin and triplet selection based on the Edit distance. We conduct a large evaluation on OnHW~\cite{ott_tist} datasets. Website: \href{https://www.iis.fraunhofer.de/de/ff/lv/dataanalytics/anwproj/schreibtrainer/onhw-dataset.html}{www.iis.fraunhofer.de/de/ff/lv/data-analytics/anwproj/schreibtrainer/onhw-dataset.html}.

The remainder of this paper is organized as follows. Section~\ref{chap_related_work} discusses related work followed by our proposed methodology in Section~\ref{chap_method}. The experimental setup is described in Section~\ref{chap_experiments} and the results are discussed in Section~\ref{chap_results}. Section~\ref{chap_conclusion} concludes.

%% file: 02related_work.tex
\section{Related Work}
\label{chap_related_work}

In this section, we address related work for OnHW recognition and for pairwise learning in relation to domain adaptation.

\paragraph{OnHW Recognition.} The novel sensor-enhanced pen based on IMUs enables new applications for writing on normal paper. First, \cite{ott} introduced a character-based dataset from sensor-enhanced pens on paper and evaluated ML and DL techniques. \cite{ott_tist} proposed several sequence-based datasets written on paper and tablet and a large benchmark of convolutional, recurrent and Transformer-based architectures, loss functions and augmentation techniques. To enhance the OnHW dataset with an offline HWR dataset, \cite{ott_ijcai} generated handwritten images with ScrabbleGAN~\cite{fogel} and improved the training with cross-modal representation learning between online and offline HWR. \cite{ott_mm} proposed a DA approach with optimal transport to adapt left-handed writers to right-handed writers for single characters. \cite{ott_wacv} reconstructed the trajectory of the pen tip for single characters written on tablet from IMU data and cameras pointing on the pen tip.

\paragraph{Pairwise Learning for DA.} Research for pairwise and triplet learning is very advanced in general \cite{chopra, schroff}, while the pairwise learning has rarely been used for sequence-based learning. \cite{guo_tang} use a triplet selection with $L_2$-normalization for language modeling. While they consier all negative pairs for triplet selection with fixed similarity intensity parameter, our triplet approach dynamically selects positive and negative samples based on ED that is closer to the temporally adaptive maximum margin function by \cite{semedo} as data is evolving over time. \texttt{DeepTripletNN}~\cite{zeng_yu_oyama} also uses the triplet loss on embeddings between time-series data (audio) and visual data. While their method uses cosine similarity for the final representation comparison, we make use of mean discrepancy and correlation techniques.

%% file: 03method.tex
\section{Methodology}
\label{chap_method}

\begin{figure*}[t!]
	\centering
    \includegraphics[width=1.0\linewidth]{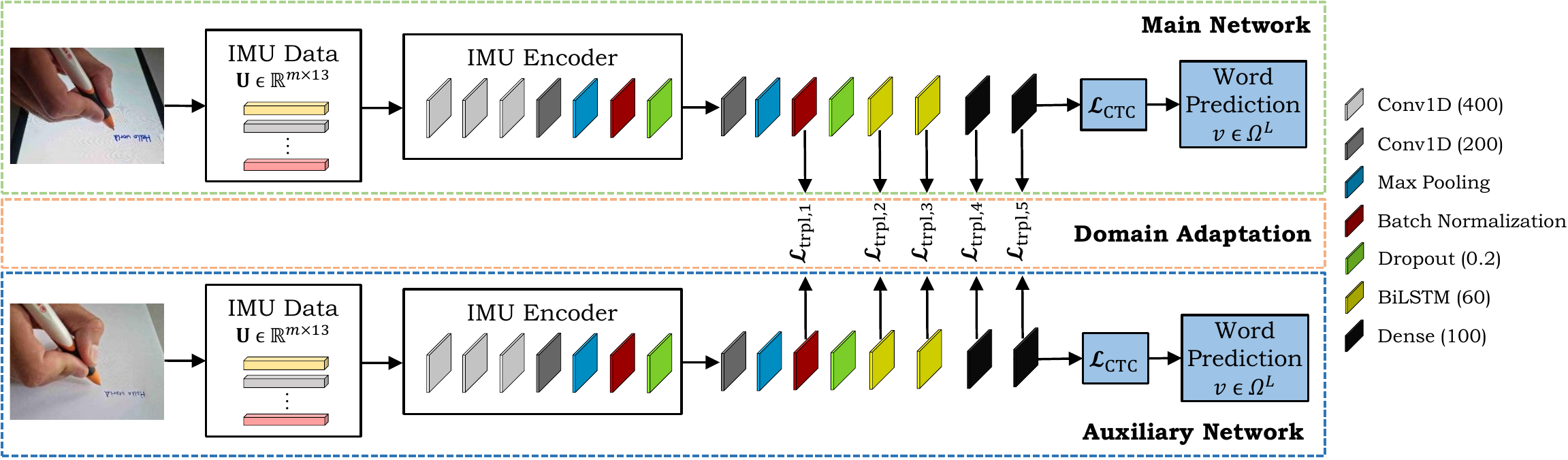}
    \caption{\textbf{Detailed method overview:} The main network (top pipeline, tablet data) and the auxiliary network (bottom pipeline, paper data) consist of the respective pre-trained architectures with convolutional and bidirectional layers. The weights are fine-tuned with domain adaptation techniques such as the triplet loss at five different fusion points.}
    \label{image_method_overview}
\end{figure*}

We start with a formal definition of multivariate time-series (MTS) classification and an method overview in Section~\ref{section_method_overview}. We propose our sequence-based triplet loss in Section~\ref{section_triplet_loss}, and finally give details about DML for DA in Section~\ref{section_da_dml}. 

\subsection{Methodology}
\label{section_method_overview}

\paragraph{MTS Classification.} We define the sensor data from pens with integrated IMUs as a MTS $\mathbf{U} = \{\mathbf{u}_1,\ldots,\mathbf{u}_m\} \in \mathbb{R}^{m \times l}$, an ordered sequence of $l \in \mathbb{N}$ streams with $\mathbf{u}_i = (u_{i,1},\ldots, u_{i,l}), i\in \{1,\ldots,m\}$, where $m \in \mathbb{N}$ is the length of the time-series. The MTS training set is a subset of the array $\mathcal{U} = \{\mathbf{U}_1,\ldots,\mathbf{U}_{n_U}\} \in \mathbb{R}^{n_U \times m \times l}$, where $n_U$ is the number of time-series. Each MTS is associated with $\mathbf{v}$, a sequence of $L$ class labels from a pre-defined label set $\Omega$ with $K$ classes. For our classification task, $\mathbf{v} \in \Omega^L$ describes words. We train a convolutional neural network (CNN) in combination with a bidirectional long short-term memory (BiLSTM). We use the connectionist temporal classification (CTC) \cite{graves} loss to predict a word $\mathbf{v}$.

\paragraph{Method Overview.} Figure~\ref{image_method_overview} gives a method overview. The \textit{main} task (top pipeline) is to classify sensor data represented as MTS with word labels $\mathbf{v}$ written with a sensor-enhanced pen \cite{ott} on tablet. The \textit{auxiliary} task (bottom pipeline) is to classify sensor data from the same sensor-enhanced pen written on paper. For optimally combining both datasets, we train a common representation between both networks by using the triplet loss $\mathcal{L}_{\text{trpl},c}$, see Section~\ref{section_triplet_loss}, with $c \in  C = \{1, 2, 3, 4, 5\}$ defines the layer both networks are combined. $c=1$ represents an intermediate fusion, while $c=5$ represents a late fusion. With DML techniques, we minimize the distance (or maximizing the similarity) between the distributions of both domains (see Section~\ref{section_da_dml}).

\subsection{Contrastive Learning and Triplet Loss}
\label{section_triplet_loss}

To learn a common representation typically pairs of same class labels of both domains are used. Pairs with similar but different labels can improve the training process. This can be achieved using the triplet loss \cite{schroff} which enforces a margin between pairs of MTS of tablet and paper sources with the same identity to all other different identities. As a consequence, the feature embedding for one and the same labels lives on a manifold, while still enforcing the distance and thus discriminability to other identities. We define the MTS $\mathbf{U}_i^a$ of the tablet dataset as \textit{anchor}, an MTS $\mathbf{U}_i^p$ of the paper dataset as the \textit{positive} sample, and an MTS $\mathbf{U}_i^n$ of the paper dataset as the \textit{negative} sample. We seek to ensure that the embedding of the anchor $f_c(\mathbf{U}_i^a)$ of a specific label is closer to the embedding of the positive sample $f_c(\mathbf{U}_i^p)$ of the same label that it is to the embedding of any negative sample $f_c(\mathbf{U}_i^n)$ of another label. Thus, we want the inequality
\begin{equation}
\label{equ_triplet1}
    \mathcal{L}_{\text{DML}} \big( f_c(\mathbf{U}_i^a), f_c(\mathbf{U}_i^p) \big) + \alpha < \mathcal{L}_{\text{DML}} \big(f_c(\mathbf{U}_i^a), f_c(\mathbf{U}_i^n) \big),
\end{equation}
to hold for all training samples $\big(f_c(\mathbf{U}_i^a), f_c(\mathbf{U}_i^p), f_c(\mathbf{U}_i^n) \big) \in \Phi$ with $\Phi$ being the set of all possible triplets in the training set. $\alpha$ is a margin between positive and negative pairs. The DML loss $\mathcal{L}_{\text{DML}}$ is defined in Section~\ref{section_da_dml}. The \textit{contrastive loss} minimizes the distance of the anchor to the positive sample and separately maximizes the distance to the negative sample. Instead, we can formulate the \textit{triplet loss} as
\begin{equation}
\label{equ_triplet2}
\begin{aligned}
    \mathcal{L}_{\text{trpl,c}}(\mathbf{U}^a, \mathbf{U}^p, \mathbf{X}^n) = \sum_{i=1}^N \max \Big[ &\mathcal{L}_{\text{DML}}\big(f_c(\mathbf{U}_i^a), f_c(\mathbf{U}_i^p)\big) - \\
    &\mathcal{L}_{\text{DML}}\big(f_c(\mathbf{U}_i^a), f_c(\mathbf{U}_i^n)\big) + \alpha, 0 \Big],
\end{aligned}
\end{equation}
\setlength{\intextsep}{6pt}
\setlength{\columnsep}{14pt}

\begin{wrapfigure}{R}{4.3cm}%
    \centering
    \includegraphics[width=1.0\linewidth]{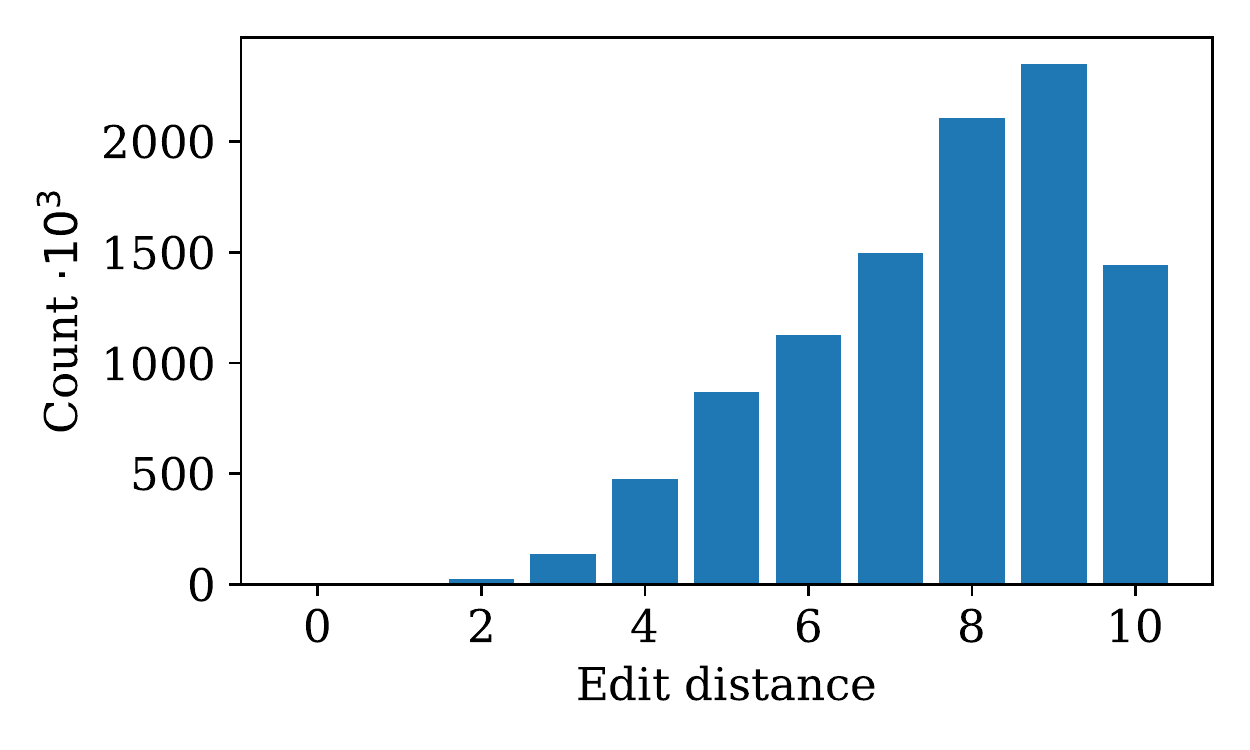}
    \caption{Number tablet-paper pairs dependent on the ED.}
    \label{image_triplet_pairs}
\end{wrapfigure}%
\noindent where $N$ is the number of triplets. To ensure fast convergence, it is necessary to select triplets that violate the constraint from Equation~\ref{equ_triplet1}. Computing the loss for all triplet pairs leads to poor training performance as poorly chosen pairs dominate hard ones \cite{do_tran}. We use the triplet selection approach by \cite{ott_ijcai} that uses the Edit distance (ED) to define the identity and select triplets. We define two sequences with an ED of 0 as positive pair, and with an ED larger than 0 as negative pair (between 1 and 10). We use only substitutions for triplet selection. Figure~\ref{image_triplet_pairs} shows the number of triplet pairs for each ED. While there exist 265 samples for $ED = 0$, 3,022 samples for $ED = 1$ and 23,983 samples for $ED = 2$, the number of pairs highly increase for higher EDs. For each batch, we search in a dictionary of negative sample pairs for samples with $ED = 1 + \lfloor\frac{\max_{e}-e-1}{20}\rfloor$ as lower bound for the current epoch $e$ and maximal epochs $\max_e = 200$ \cite{ott_ijcai}. For every pair we randomly select one paper sample. We let the margin $\alpha$ in the triplet loss vary for each batch such that $\alpha = \beta \cdot \overline{ED}$ is depending on the mean ED of the batch and is in the range $[1, 11]$. $\beta$ depends on the DML loss (see Section~\ref{section_da_dml}).

\subsection{Domain Adaptation with Deep Metric Learning}
\label{section_da_dml}

A domain $\mathcal{D}$ consists of a feature space $\mathcal{X}$ with marginal probability $P(\mathcal{X})$. The task is defined by the label space $\mathcal{Y}$. When considering OnHW recognition, there is a source domain (paper dataset) $\mathcal{D}_S = \{\mathcal{U}_S^i, \mathcal{Y}_S^i\}_{i=1}^{\mathcal{N}_S}$ of $\mathcal{N}_S$ labeled samples of $|\mathcal{Y}^i_S|$ categories, and a target domain (tablet dataset) $\mathcal{D}_T = \{\mathcal{U}_T^i, \mathcal{Y}_T^i\}_{i=1}^{\mathcal{N}_T}$ of $\mathcal{N}_T$ labeled samples of $|\mathcal{Y}_T^i|$ categories. DA can mitigate the domain shift and improve the classification accuracy in the target domain by enforcing the distance of target embeddings $f_c(\mathbf{U}_i^a)$ and source domain embeddings $f_c(\mathbf{U}_i^p)$ and $f_c(\mathbf{U}_i^n)$ to be minimal. The embeddings are of size $400 \times 200$ for $c=1$, of size $60 \times 200$ for $c=2$ and $c=3$, and of size $60 \times 100$ for $c=4$ and $c=5$. We search for a DML loss $\mathcal{L}_{\text{DML}}\big(f_c(\mathbf{U}_i^a), f_c(\mathbf{U}_i^p)\big)$, respectively for the negative sample $\mathbf{U}_i^n$, that takes the domain shift into account. To perform domain alignment, we use higher-order moment matching (HoMM) \cite{chen_fu}
\begin{equation}
\label{equ_homm}
    \mathcal{L}_{\text{HoMM}}\big(f_c(\mathbf{U}_i^a), f_c(\mathbf{U}_i^p)\big) = \frac{1}{H^p} \Bigg|\Bigg| \frac{1}{n_s} \sum_{i=1}^{n_s} f_c(\mathbf{U}_i^a)^{\otimes p} - \frac{1}{n_t} \sum_{i=1}^{n_t} f_c(\mathbf{U}_i^p)^{\otimes p} \Bigg|\Bigg|_F^2,
\end{equation}
between embeddings $f_c(\mathbf{U}_i^a)$ and $f_c(\mathbf{U}_i^p)$, respectively for $f_c(\mathbf{U}_i^a)$ and $f_c(\mathbf{U}_i^n)$. It holds $n_s = n_t = b$ with the batch size $b$. $||\cdot||_F$ denotes the Frobenius norm, $H$ is the number of hidden neurons in the adapted layer, and $(\cdot)^{\otimes p}$ denotes the $p$-level tensor power. When $p=1$, HoMM is equivalent to the linear MMD~\cite{tzeng}, and when $p=2$, HoMM is equivalent to the Gram matrix matching. When the embeddings are normalized by subtracting the mean, the centralized Gram matrix turns into the covariance matrix \cite{chen_fu}, and hence, HoMM for $p=2$ is equivalent to CORAL~\cite{sun_saenko}. However, the space complexity for calculating the tensor $(\cdot)^{\otimes p}$ reaches $\mathcal{O}(H^p)$. This can be reduced by \textit{group moment matching} that divides the hidden neurons into $n_g$ groups, with each group $\lfloor\frac{H}{n_g}\rfloor$ neurons, and the space complexity reduces to $\mathcal{O}(n_g \cdot \lfloor\frac{H}{n_g}\rfloor^p)$. Furthermore, \textit{random sampling matching} randomly selects $T$ values in the high-level tensor, and only aligns these $T$ values in the source and target domains. The space complexity reduces to $\mathcal{O}(T)$ \cite{chen_fu}. For our application, we evaluate orders $p=1$, $p=2$ and $p=3$, and choose $T=1,000$, which reaches the limits of our training setup of GPUs with 32 GB VRAM. Alternatively, we make use of (Jeff and Stein) CORAL~\cite{sun_saenko}. We choose the hyperparameters $\beta$ from Section~\ref{section_triplet_loss} proposed in Table~\ref{table_hyper_beta}.

\begin{table}[t!]
\begin{center}
\setlength{\tabcolsep}{4.0pt}
    \caption{Hyperparameter choices of $\beta$ for all DML loss functions and fusion points $c$.}
    \label{table_hyper_beta}
    \begin{tabular}{ p{0.5cm} | p{0.5cm} | p{0.5cm} | p{0.5cm} | p{0.5cm} | p{0.5cm} }
    \multicolumn{1}{c|}{\textbf{DA Loss}} &  \multicolumn{1}{c|}{$c=1$} &  \multicolumn{1}{c|}{$c=2$} &  \multicolumn{1}{c|}{$c=3$} &  \multicolumn{1}{c|}{$c=4$} &  \multicolumn{1}{c}{$c=5$} \\ \hline
    \multicolumn{1}{l|}{kMMD~\cite{long_zhu_mmd} ($p=1$)} &  \multicolumn{1}{r|}{10} &  \multicolumn{1}{r|}{100} &  \multicolumn{1}{r|}{100} &  \multicolumn{1}{r|}{10} &  \multicolumn{1}{r}{10} \\
    \multicolumn{1}{l|}{HoMM~\cite{chen_fu} ($p=2$)} &  \multicolumn{1}{r|}{0.01} &  \multicolumn{1}{r|}{10\textsuperscript{5}} &  \multicolumn{1}{r|}{10\textsuperscript{4}} &  \multicolumn{1}{r|}{100} &  \multicolumn{1}{r}{0.1} \\
    \multicolumn{1}{l|}{HoMM~\cite{chen_fu} ($p=3$)} &  \multicolumn{1}{r|}{10\textsuperscript{-6}} &  \multicolumn{1}{r|}{10\textsuperscript{6}} &  \multicolumn{1}{r|}{10\textsuperscript{5}} &  \multicolumn{1}{r|}{100} &  \multicolumn{1}{r}{10\textsuperscript{-3}} \\
    \multicolumn{1}{l|}{kHoMM~\cite{chen_fu} ($p=2$)} &  \multicolumn{1}{r|}{10\textsuperscript{3}} &  \multicolumn{1}{r|}{10\textsuperscript{6}} &  \multicolumn{1}{r|}{10\textsuperscript{6}} &  \multicolumn{1}{r|}{10\textsuperscript{4}} &  \multicolumn{1}{r}{10} \\
    \multicolumn{1}{l|}{kHoMM~\cite{chen_fu} ($p=3$)} &  \multicolumn{1}{r|}{100} &  \multicolumn{1}{r|}{10\textsuperscript{6}} &  \multicolumn{1}{r|}{10\textsuperscript{6}} &  \multicolumn{1}{r|}{10\textsuperscript{4}} &  \multicolumn{1}{r}{10} \\
    \multicolumn{1}{l|}{CORAL~\cite{sun_feng}} &  \multicolumn{1}{r|}{0.01} &  \multicolumn{1}{r|}{10\textsuperscript{4}} &  \multicolumn{1}{r|}{10\textsuperscript{4}} &  \multicolumn{1}{r|}{10} &  \multicolumn{1}{r}{0.01} \\
    \multicolumn{1}{l|}{Jeff CORAL~\cite{sun_feng}} &  \multicolumn{1}{r|}{0.1} &  \multicolumn{1}{r|}{100} &  \multicolumn{1}{r|}{100} &  \multicolumn{1}{r|}{1} &  \multicolumn{1}{r}{0.1} \\
    \multicolumn{1}{l|}{Stein CORAL~\cite{sun_feng}} &  \multicolumn{1}{r|}{1} &  \multicolumn{1}{r|}{100} &  \multicolumn{1}{r|}{100} &  \multicolumn{1}{r|}{10} &  \multicolumn{1}{r}{1} \\
    \end{tabular}
\end{center}
\end{table}

%% file: 04experiments.tex
\section{Experiments}
\label{chap_experiments}

\textbf{OnHW recognition} uses time in association with different types of spatio-temporal signal. The pen in \cite{ott} uses two accelerometers (3 axes each), one gyroscope (3 axes), one magnetometer (3 axes), and one force sensor at 100\,Hz. One sample of size $m \times l$ represents an MTS of $m$ time steps from $l=13$ sensor channels. We make use of three sequence-based datasets proposed by \cite{ott_tist}: The \textit{OnHW-words500} dataset contains 500 repeated words from 53 writers. The \textit{OnHW-wordsRandom} contains randomly selected words from 54 writers. Both datasets combined represent the (auxiliary task) dataset from source domain written on paper, and contains in total 39,863 samples. The \textit{OnHW-wordsTraj} dataset contains 4,262 samples of randomly selected words from two writers, and represents the (main task) dataset from target domain written on tablet. The challenging task is to adapt on one of the two writers (who collected data on tablet) by utilizing the paper datasets. We make use of 80/20 train/validation splits for writer-dependent (WD) and writer-independent (WI) evaluation.

\textbf{Language Models (LMs).} We apply LMs to the softmax output values of the neural networks. We use the Wikimedia database by the Wikimedia Foundation \cite{wikimedia}. We create the n-gram dictionaries with the \texttt{nltk} package \cite{nltk} and exclude punctuation marks. These dictionaries store the probabilities of the order of characters generated from sequences of items. Next, we select the paths (word length $\times$ number of character labels) of the network predictions with the highest softmax values with a softmax threshold of 0.001. For more than $path\_thresh = 512$ available paths, we limit the number of paths to $max\_paths = 50$. Lastly, the n-gram models are applied to these paths.

%% file: 05evaluation.tex
\section{Experimental Results}
\label{chap_results}

\textit{Hardware and Training Setup.} For all experiments we use Nvidia Tesla V100-SXM2 GPUs with 32 GB VRAM equipped with Core Xeon CPUs and 192 GB RAM. We use the vanilla Adam optimizer with a learning rate of $10^{-4}$. We pre-train both networks for 1,000 epochs, and adapt for 200 epochs for the contrastive loss and 2,000 epochs for the triplet loss. A metric for sequence evaluation is the character error rate (CER) and the word error rate (WER) defined through the ED (see \cite{ott_tist}).

\subsection{Baseline Results} 
\label{section_eval_baseline}

\begin{figure}[t!]
	\centering
    \includegraphics[trim=10 10 10 10, clip, width=1.0\linewidth]{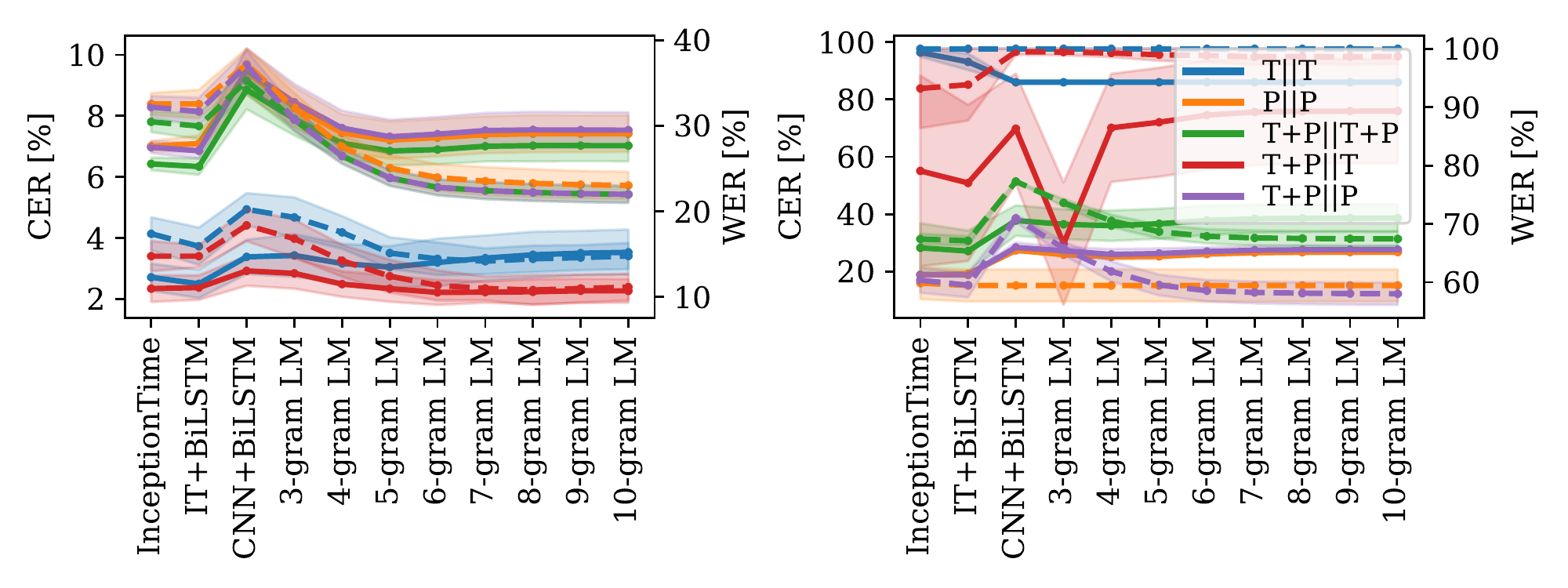}
    \caption{Baseline results (WER: dashed, CER: solid, in \%, averaged over cross-validation splits) for the InceptionTime (IT) \cite{fawaz} and our CNN+BiLSTM architectures and different n-gram LMs. Left: WD task. Right: WI task. Legend (tablet = T, paper = P): First notes training set and second notes validation set of the OnHW datasets \cite{ott_tist}.}
    \label{image_baseline_results}
\end{figure}

Figure~\ref{image_baseline_results} presents baseline results for our CNN+BiLSTM architecture compared to InceptionTime (IT) \cite{fawaz} with and without additional BiLSTM layer. Consistently, IT+Bi-LSTM outperforms IT, while the CER and WER slightly increases for our CNN+Bi-LSTM model. For all datasets, the WD classification task shows better performance than the WI task. We can improve the CNN+BiLSTM results with the LM from 3.38\% to 3.04\% CER and from 20.23\% to 15.13\% WER with 5-gram LM trained on the tablet dataset. While the WER consistently decreases with higher n-gram LM, the CER increases higher than 5-gram LM. This is more notable for the separate tablet (T) dataset as the length of the words are here shorter than for the paper (P) datasets. By simply combining both datasets, the models achieve lower error rates evaluated on the tablet dataset only (from 3.04\% CER for T$\|$T to 2.34\% CER for T+P$\|$T for 5-gram LM), but increases for the model evaluated on the paper dataset only (from 7.21\% CER for P$\|$P to 7.32\% CER for T+P$\|$P for 5-gram LM). We define X$\|$Y, with X notes training dataset and Y notes validation dataset. This demonstrates the problem that there is a domain shift between tablet and paper data and that the size of the tablet dataset is small.

\subsection{Domain Adaptation Results} 
\label{section_eval_da}

\begin{figure}[t!]
	\centering
	\begin{minipage}[b]{0.485\linewidth}
        \centering
        \includegraphics[trim=1 11 1 10, clip, width=1.0\linewidth]{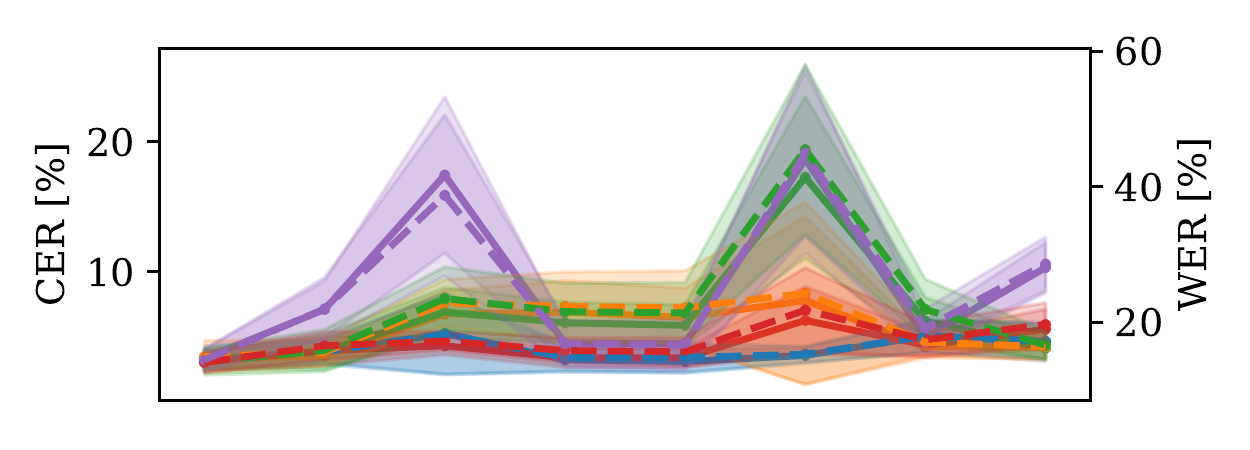}
        \subcaption{Contrastive loss, validation: tablet, WD.}
        \label{image_CER_C_T_WD}
    \end{minipage}
    \hfill
	\begin{minipage}[b]{0.485\linewidth}
        \centering
        \includegraphics[trim=9 11 10 10, clip, width=1.0\linewidth]{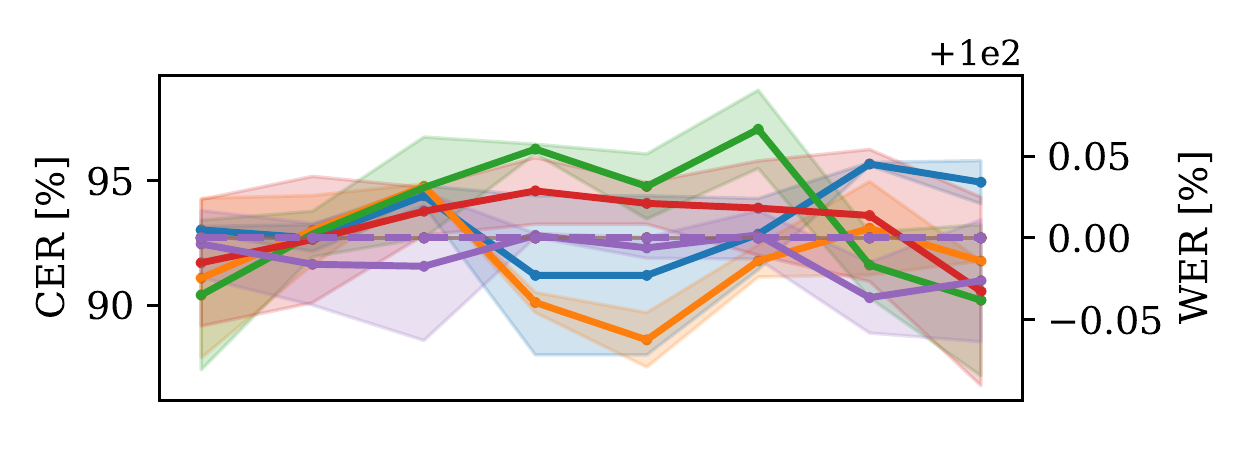}
        \subcaption{Contrastive loss, validation: tablet, WI.}
        \label{image_CER_C_T_WI}
    \end{minipage}
    \hfill
	\begin{minipage}[b]{0.485\linewidth}
        \centering
        \includegraphics[trim=1 11 1 10, clip, width=1.0\linewidth]{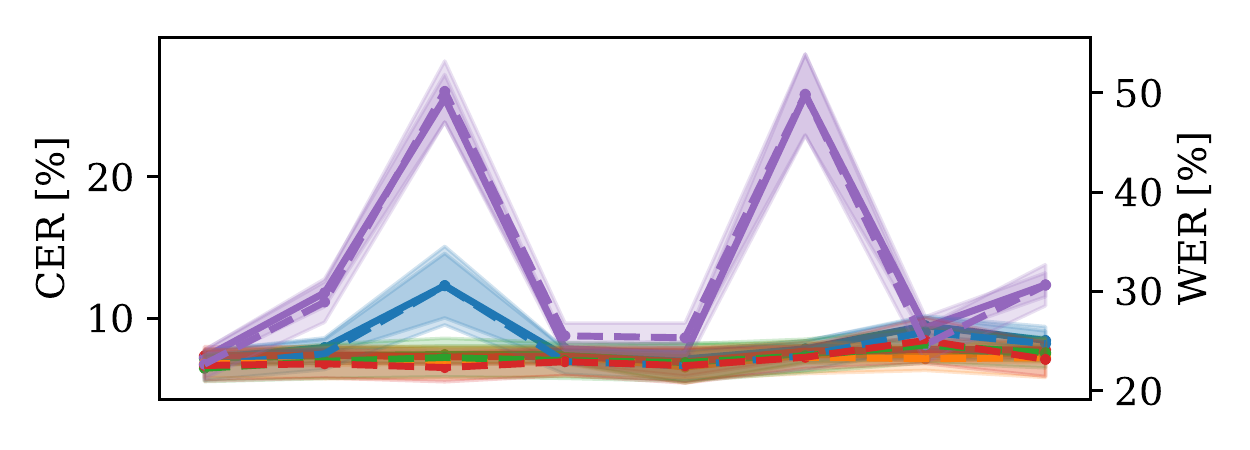}
        \subcaption{Contrastive loss, validation: paper, WD.}
        \label{image_CER_C_P_WD}
    \end{minipage}
    \hfill
	\begin{minipage}[b]{0.485\linewidth}
        \centering
        \includegraphics[trim=14 11 1 10, clip, width=1.0\linewidth]{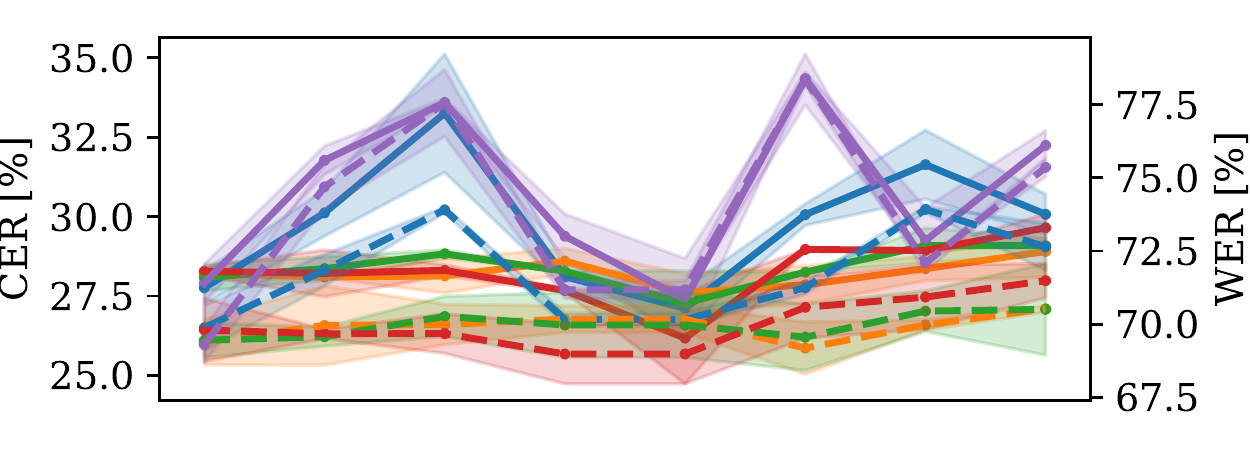}
        \subcaption{Contrastive loss, validation: paper, WI.}
        \label{image_CER_C_P_WI}
    \end{minipage}
    \hfill
	\begin{minipage}[b]{0.485\linewidth}
        \centering
        \includegraphics[trim=1 11 1 10, clip, width=1.0\linewidth]{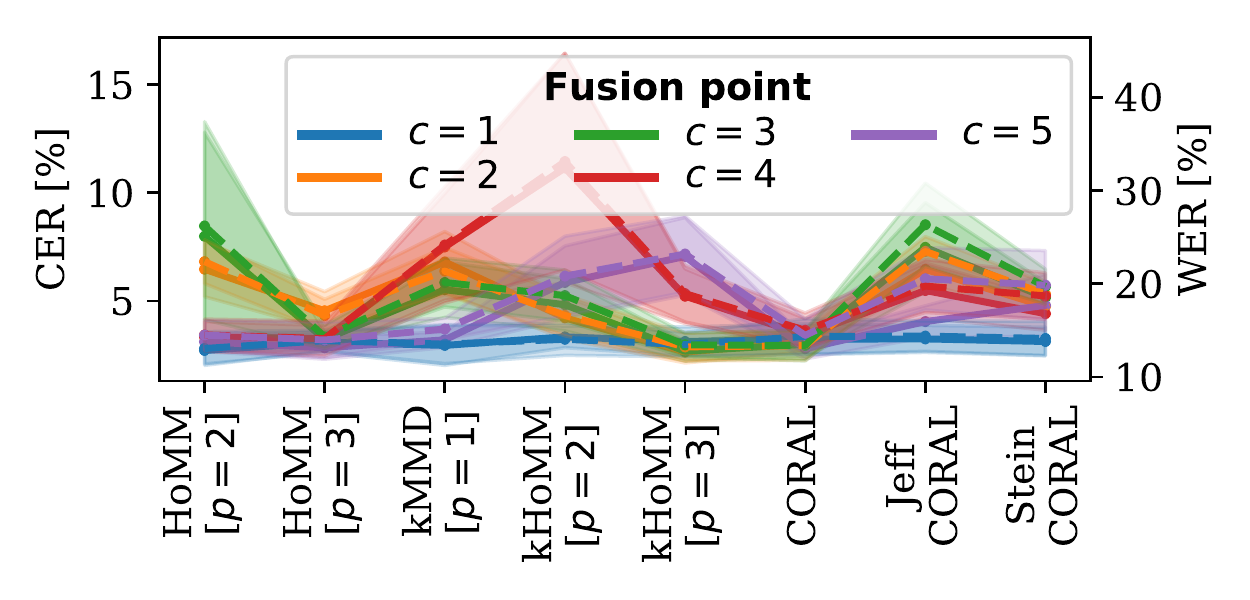}
        \subcaption{Triplet loss, validation: tablet, WD.}
        \label{image_CER_T_T_WD}
    \end{minipage}
    \hfill
	\begin{minipage}[b]{0.485\linewidth}
        \centering
        \includegraphics[trim=9 11 10 10, clip, width=1.0\linewidth]{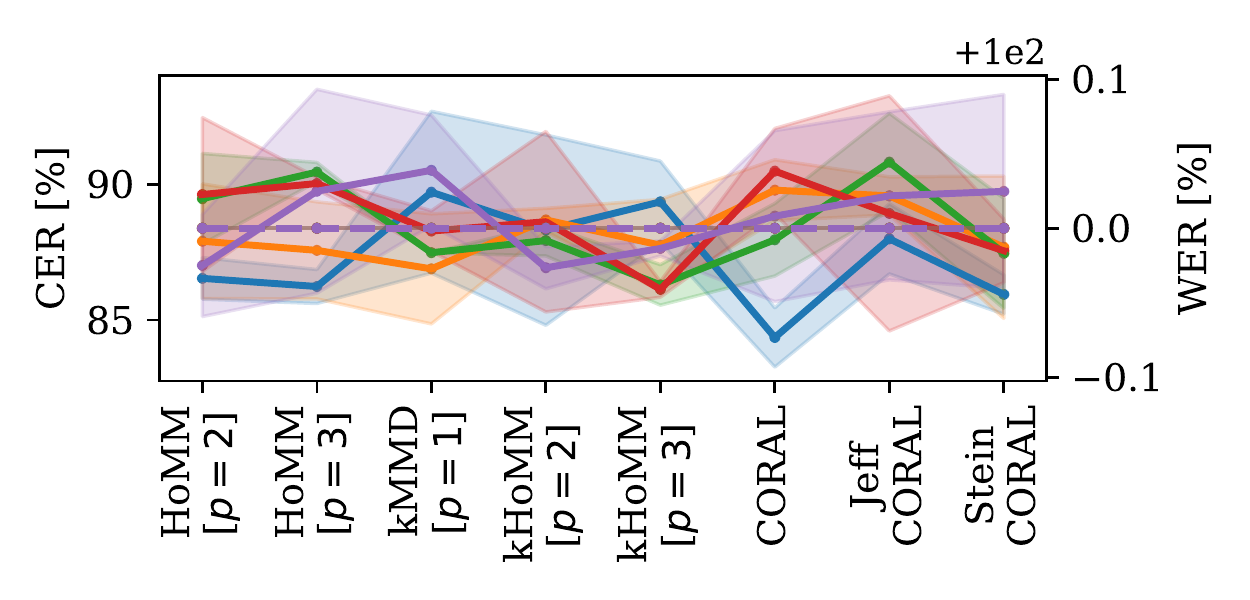}
        \subcaption{Triplet loss, validation: tablet, WI.}
        \label{image_CER_T_T_WI}
    \end{minipage}
    \caption{DA results (WER: dashed, CER: solid, in \%, averaged over cross-validation splits) for our CNN+BiLSTM architecture (with a 6-gram LM for tablet trainings, 10-gram for paper trainings respectively, for WD tasks and without LMs for WI tasks). The model is trained with the combined tablet and paper dataset with different representation loss functions at five different fusion points $c$.}
    \label{image_da_results_all}
\end{figure}

We train the contrastive and pairwise learning approach by adapting paper data to tablet data with different representation loss functions (HoMM~\cite{chen_fu} and CORAL~\cite{sun_saenko}) and propose results in Figure~\ref{image_da_results_all}. State-of-the-art pairwise learning techniques cannot be applied to our setup as they are typically proposed for single label classification tasks. While the contrastive loss cannot improve the tablet validation results (\ref{image_CER_C_T_WD}), the validation on paper (\ref{image_CER_C_P_WD}) does improve (orange and purple curve of Figure~\ref{image_baseline_results}). Also for the WI task, the paper validation improves (\ref{image_CER_C_P_WI}), while the tablet dataset is still a challenging task (\ref{image_CER_C_T_WI}). The triplet loss is on par with the baseline results for the WD task (\ref{image_CER_T_T_WD}). We see that early fusion ($c=1$) leads to consistently low CERs as it is possible to adapt more trainable parameters after this fusion point. Intermediate ($c=2$ and $c=3$) and late ($c=4$ and $c=5$) fusion is dependent on the representation loss. kHoMM of order $p=3$ at $c=3$ leads to the highest significant improvement of 13.45\% WER and 2.68\% CER. The error rates of Jeff and Stein CORAL are marginally higher. LMs for $c=4$ and $c=5$ decrease results as the softmax output values are lower (uncertainty is higher) and the LM often chooses the second largest softmax value. We summarize the main difficulties as following: (1) While the paper dataset is very large, the tablet dataset as target domain is rather small. This leads to a small number of pairs with a small ED (see Figure~\ref{image_triplet_pairs}). (2) Furthermore, as the OnHW-words500 dataset contains the same 500 word labels per writer, the variance of specific positive pairs is small.

%% file: 06conclusion.tex
\section{Conclusion}
\label{chap_conclusion}

We proposed a DA approach for online handwriting recognition for sequence-based classification tasks between writing on tablet and paper. For this, contrastive and triplet losses enhance the main dataset and allows a more generalized training process. We evaluated moment matching techniques as DML loss functions. The kernalized HoMM of order 3 at the intermediate fusion layer combined with a n-gram language model provides the lowest error rates.